\documentclass{article}

\usepackage{arxiv}

\usepackage[utf8]{inputenc} 
\usepackage[T1]{fontenc}    
\usepackage{hyperref}       
\usepackage{url}            
\usepackage{booktabs}       
\usepackage{amsfonts}       
\usepackage{nicefrac}       
\usepackage{microtype}      
\usepackage{cleveref}       
\usepackage{lipsum}         
\usepackage{tabularx}       
\usepackage{graphicx}
\usepackage[numbers]{natbib}
\usepackage{doi}

\title{Effectiveness of Zero-shot-CoT in Japanese Prompts}

\date{}

\author{
    \href{https://orcid.org/0009-0003-9732-297X}{Shusuke Takayama} \\
    Future University Hakodate \\
    \texttt{b1020117@gmail.com} \\
    \\
    \href{https://orcid.org/0009-0001-3619-4121}{Ian Frank} \\
    Future University Hakodate \\
    \texttt{ianf@fun.ac.jp}
}

\begin{document}

\maketitle

\begin{abstract}

We compare the effectiveness of zero-shot Chain-of-Thought (CoT) prompting in Japanese and English using ChatGPT-3.5 and 4o-mini.  The technique of zero-shot CoT, which involves appending a phrase such as ``Let’s think step by step'' to a prompt to encourage reasoning before answering, has been shown to offer LLM performance improvements in mathematical and reasoning tasks, particularly in English. We investigate how these effects transfer to Japanese using the Japanese Multi-task Language Understanding Benchmark (JMMLU) and the Multi-task Language Understanding Benchmark (MMLU).  Our results show that while zero-shot CoT prompting can lead to notable performance gains for some prompt categories in GPT-3.5, its impact in GPT-4o-mini is associated with significant performance declines.  However, for Japanese prompts there remain certain categories, such as college mathematics and abstract algebra, that still exhibit improvements, despite the broader trend of diminishing effectiveness in more advanced models.  \end{abstract}

\keywords{Chain of Thought (CoT), LLMs, Prompt Engineering.}

\section{Introduction}

This study addresses issues of prompt engineering in Japanese by examining the effectiveness of zero-shot CoT prompting in GPT-3.5 and GPT-4o-mini.  Prompt engineering plays a crucial role in optimising large language models (LLMs) for different tasks~\cite{radford2019gpt2, brown2020fewshot}. While the impact varies across domains, structured prompting strategies have been shown to improve reasoning-intensive tasks such as mathematics and logical inference. However, the effectiveness of these techniques can differ depending on the model architecture, language, and task type. Existing studies have a primary focus on English prompts, with other languages such as Japanese being relatively under-explored.  As one way to address this disparity, we evaluate zero-shot CoT across multiple domains using the Multi-task Language Understanding Benchmark (MMLU)~\cite{hendrycks2021MMLU} and its Japanese counterpart, the JMMLU~\cite{jmmlu2023}. By analysing model performance on these benchmarks, we aim to shed light on how the effectiveness of prompting varies across languages.

\section{Related Research}

Many studies have explored the factors influencing LLM performance, such as prompting techniques, reasoning capabilities, emotional context, and prompt sensitivity. The use of the Chain-of-Thought (CoT) terminology in prompting originated in~\cite{wei2022chain}, where it was shown to improve performance in complex reasoning tasks by encouraging step-by-step breakdowns of problems.  For example, in evaluations on the GSM8K math reasoning dataset, text-davinci-002 achieved a zero-shot score of 25.5 ± 2.1, but applying zero-shot-CoT improved this to 64.1 ± 1.8. Adding six additional reasoning steps further increased the score to 78.8 ± 0.3.

Zero-shot-CoT is a simplified form of CoT prompting that involves appending a phrase such as ``Let’s think step by step'' to prompts. It has been found to improve performance in multi-step reasoning tasks such as arithmetic problems~\cite{kojima2023large}.

Overall, CoT can be considered a specialised form of in-context learning (ICL)~\cite{brown2020fewshot} that focuses on reasoning rather than task adaptation. For LLMs such as GPT-3.5 and GPT-4 that are the focus of this paper, research has shown a linear relationship between the number of reasoning steps and accuracy, with Palm-540B achieving notable gains across datasets like GSM8K and SVAMP~\cite{jin2024impact}.

Newer models such as o1, with their apparent integration of CoT-like reasoning internally, may implicitly blur the distinction between ICL-driven CoT and built-in model behavior, potentially reducing the need for explicit prompting. However, a study spanning SOTA models, including o1-mini, used minor rewordings or additions of irrelevant clauses to the GSM8K dataset to reveal that LLMs struggle with true mathematical understanding~\cite{mirzadeh2024gsm}. The study concluded that LLM performance was ``fragile'' and ``may resemble sophisticated pattern matching more than true logical reasoning''.

As a further tool to investigate these limitations, a dataset derived from the William Lowell Putnam Mathematical Competition was introduced by~\cite{gulati2024putnam}. Results showed that the programmatic alteration of problem elements like variables and constants caused most LLMs to perform significantly worse, indicating a reliance on memorised patterns rather than true mathematical reasoning.

Effects of phrasing and emotion in prompts have also been investigated.  For example,~\cite{yin2024should} compared politeness in three languages, finding that in Japanese, less polite prompts tended to yield better performance,
whereas in Chinese, increased politeness generally improved performance, and in English, GPT-3.5 performed best with highly polite prompts.  These findings suggest that excessive politeness does not necessarily enhance model performance and that prompt effectiveness may be culturally or linguistically dependent.  Another study examined the impact of emotional language in prompts, finding that adding encouraging phrases such as ``You can do it!'' improved task performance~\cite{li2023large}. Across multiple models, prompts with emotional expressions consistently outperformed both original and zero-shot-CoT prompts, particularly in human-evaluated tasks like poetry composition and summarisation. This suggests that beyond structured reasoning, motivational and affective elements in prompts can impact model performance.

Overall, the research landscape highlights the evolving role of explicit prompting techniques as LLMs become more capable of performing structured reasoning without external guidance. As newer models increasingly integrate implicit CoT reasoning, further research will be needed to refine effective prompting strategies, and to explore how insights from different languages can inform broader improvements in model behaviour.

\section{Evaluation Study}

\begin{figure}[htb]
	\centering  
        \includegraphics[width=0.33\textwidth]{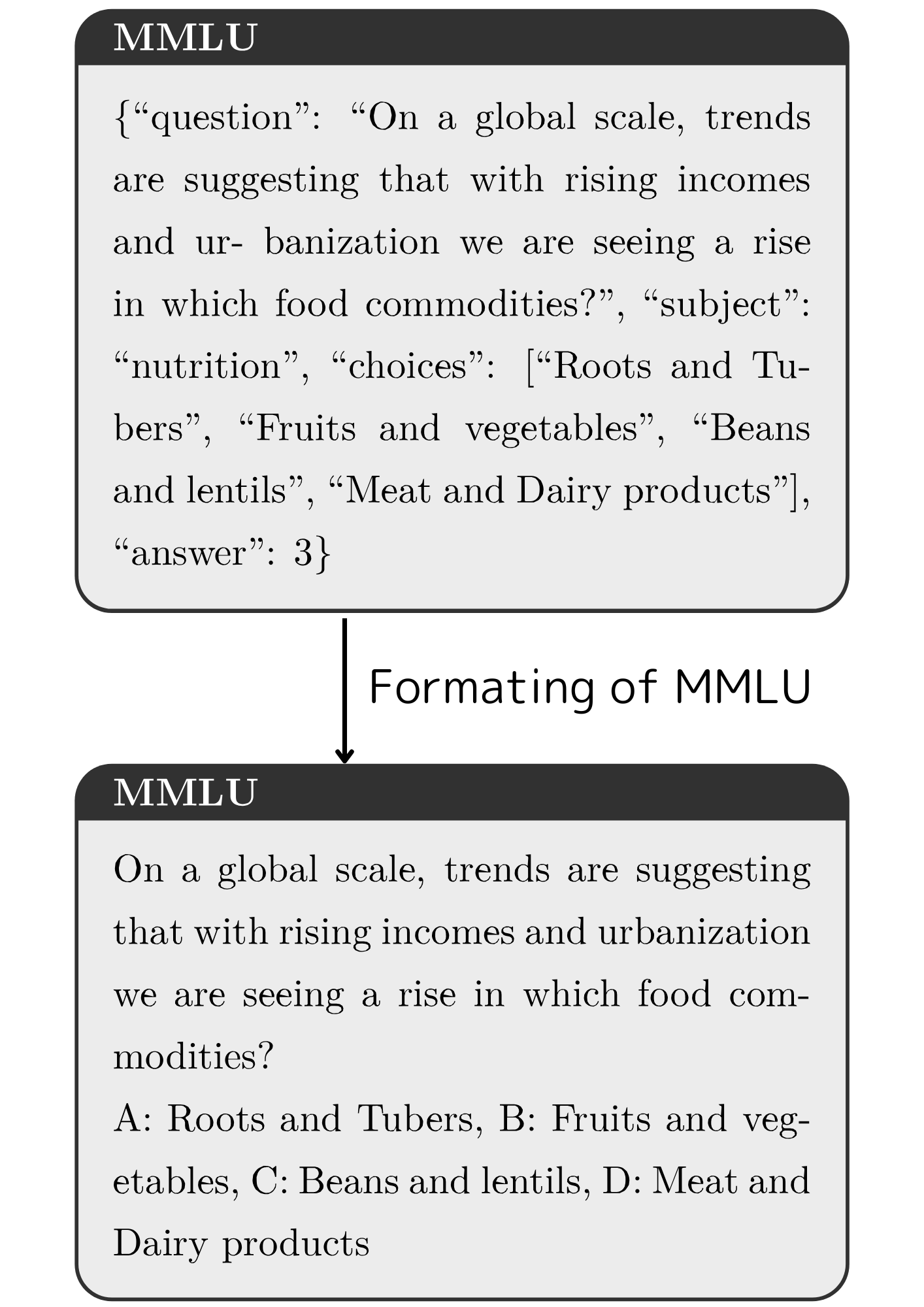}
        \includegraphics[width=0.33\textwidth]{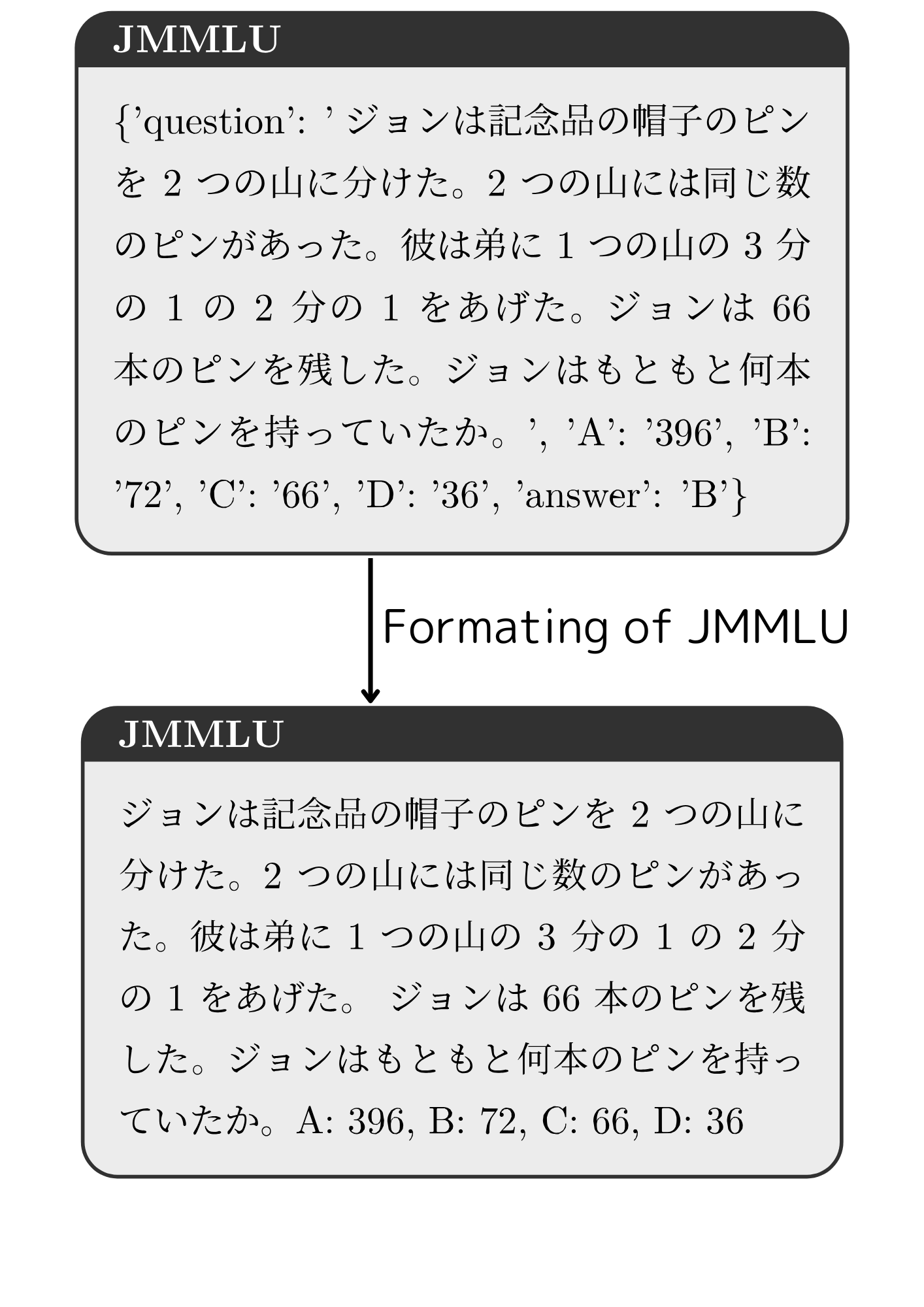}
	\caption{Example of question formatting in MMLU and JMMLU, allowing evaluation under comparable conditions.}
	\label{fig:fig1}
\end{figure}

JMMLU is a Japanese translation of some tasks from the multi-task language understanding benchmark MMLU, and both datasets are formatted as multiple-choice questions with four options. To use JMMLU and MMLU as evaluation datasets, the ``answer'' text was removed, and the possible options
were made distinguishable by the letters A, B, C, and D, as shown in Figure~\ref{fig:fig1}. 

All tasks in JMMLU were executed, whereas for MMLU, experimental costs were reduced by selecting the first 150 instances for each subject, with the number 150 chosen to correspond to the maximum number of instances in the JMMLU datasets. Around half of the JMMLU subjects have 150 tasks: for completeness, we list those with fewer tasks in Table~\ref{tab:JMMLU-numbers}.

\begin{table}[tbhp]
    \caption{JMMLU subjects with fewer than 150 tasks. Subjects not listed have 150 tasks.}
    \label{tab:JMMLU-numbers}
    \centering
    \renewcommand{\arraystretch}{1.2} 
    \begin{tabularx}{\linewidth}{X c | X c} 
    \hline
    \textbf{Subject} & \textbf{Number of Tasks} & \textbf{Subject} & \textbf{Number of Tasks} \\
    \hline\hline
    nutrition & 149 &  econometrics &113\\   
     high school microeconomics & 149 & machine learning&111\\ %
     high school chemistry & 149 &  public relations&109\\
     high school macroeconomics & 148 & jurisprudence&107\\
     moral disputes & 148 & management&102\\
     astronomy & 148 & college physics & 100 \\
    high school biology & 148 & medical genetics & 99 \\
     world religions & 147 & college computer science & 99 \\
     electrical engineering & 144 & college mathematics & 99 \\
     college biology & 143 & abstract algebra & 99 \\
     japanese geography & 139 & computer security & 99 \\
    anatomy & 132 & college chemistry & 99 \\
    human sexuality & 130 &  high school computer science & 98 \\
    formal logic & 125 & global facts & 97 \\
    international law & 120 & business ethics & 86 \\
    \end{tabularx}
  \end{table}

  For the prompts, scores were measured with Japanese prompts given for JMMLU and English prompts given for MMLU, both with and without the zero-shot-CoT the sentence ``Let’s think step by step'' at the end of the tasks (either in English or Japanese). The letter A, B, C or D was extracted from the generated text to measure scores.
The overall performance is summarised in Table~\ref{tab:overall}, showing that,
on average, CoT has a negative effect across all tested tasks in both languages.

\begin{table}[htbp]
    \caption{Overall performance comparison of GPT-3.5 and GPT-4o-mini with and without zero-shot CoT, showing a general decline in performance despite some task-level improvements.}
    \centering 
    \renewcommand{\arraystretch}{1.5} 
    \begin{tabular}{p{2cm}ccc|ccc}
    \hline\hline 
        & \multicolumn{3}{c|}{\bf Japanese} & \multicolumn{3}{c}{\bf English} \\
        & CoT & no CoT & CoT change & CoT & no CoT & CoT change \\
    \hline\hline 
        GPT3.5 & 0.469 & 0.528 & -0.059 & 0.580 & 0.602 & -0.022 \\ 
    \hline 
        GPT4o-mini & 0.332 & 0.666 & -0.334 & 0.258 & 0.682 & -0.424 \\ 
    \hline 
    \end{tabular}
    \label{tab:overall}
\end{table}

\begin{figure}[p]
	\centering
        \includegraphics[width=1.0\textwidth]{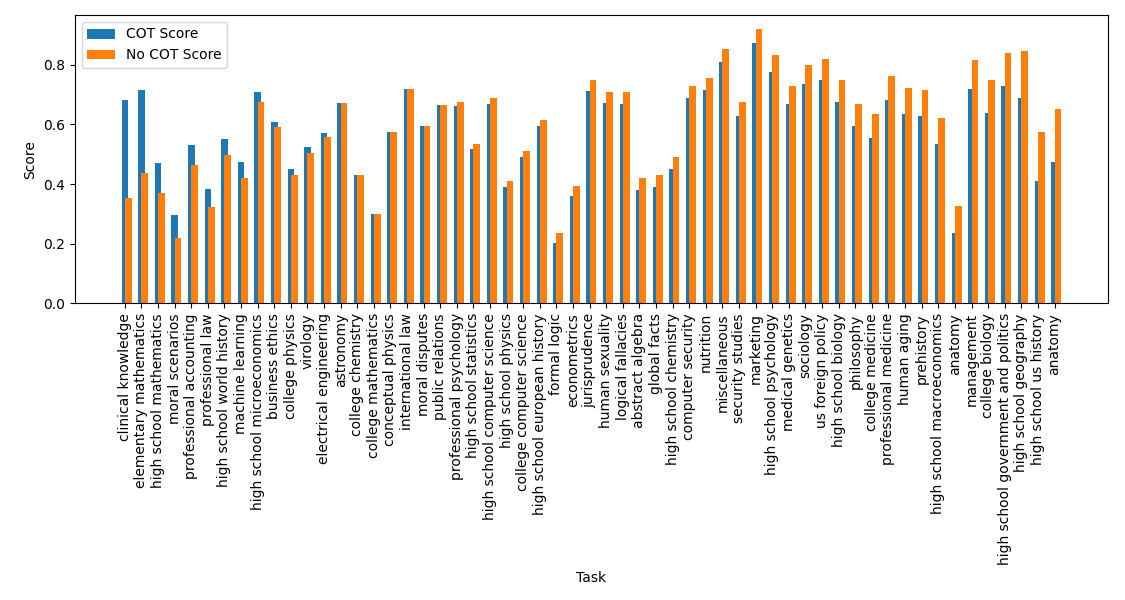}
	\caption{Performance across subjects for GPT-3.5 in English, comparing results with and without zero-shot CoT. While CoT improves performance in some reasoning-based subjects, the overall impact is mixed.}
	\label{fig:gpt3E}
\end{figure}

\begin{figure}[p]
	\centering
        \includegraphics[width=1.0\textwidth]{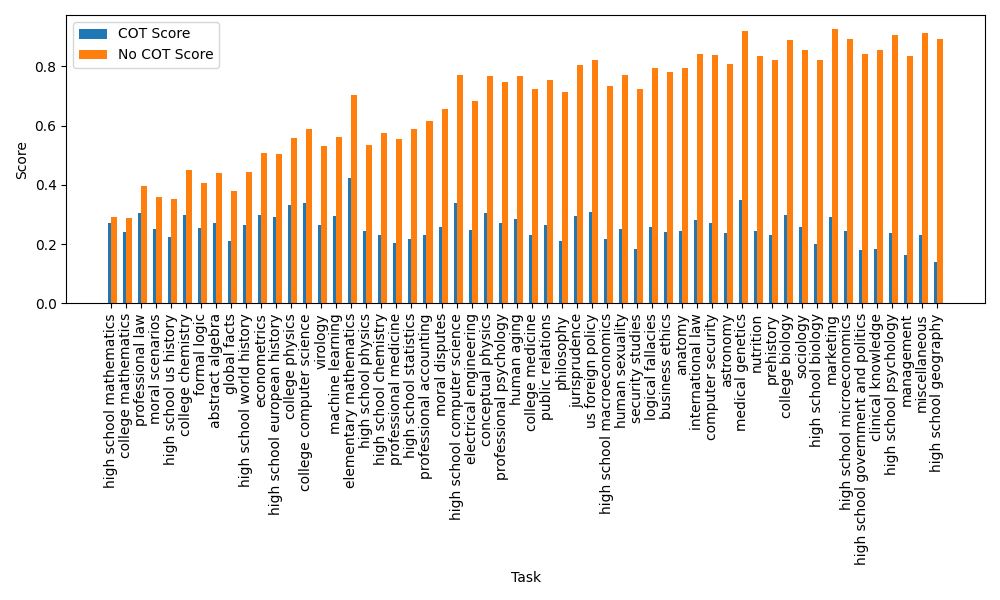}
	\caption{Performance across subjects for GPT-4o-mini in English. Unlike GPT-3.5, zero-shot CoT leads to a consistent decline, with no subjects showing improvement.}
	\label{fig:gtp4E}
\end{figure}

\begin{figure}[p]
	\centering
        \includegraphics[width=1.0\textwidth]{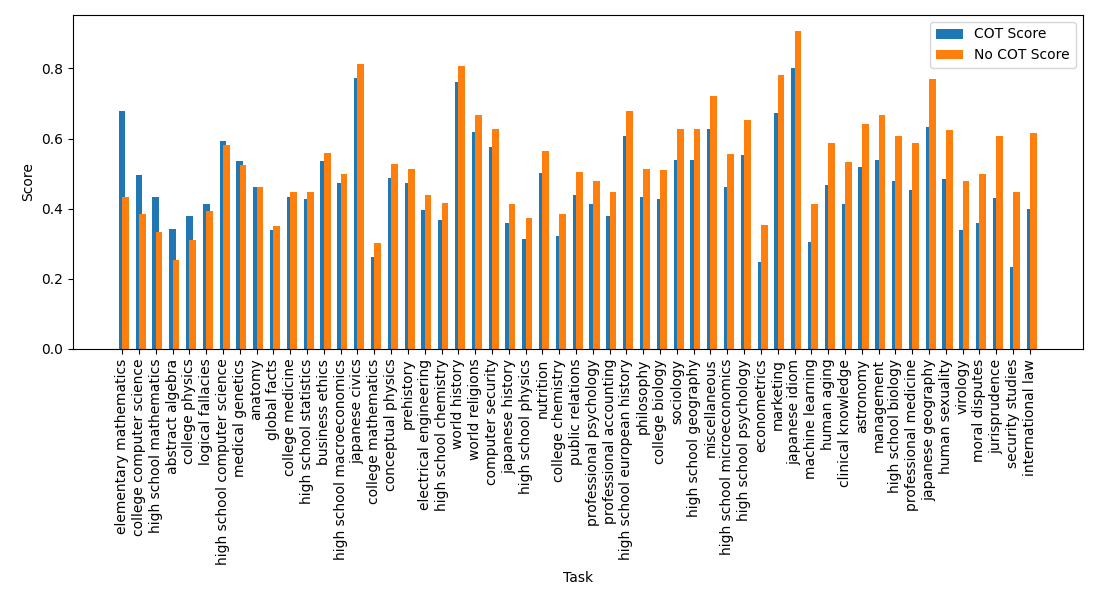}
	\caption{Performance across subjects for GPT-3.5 in Japanese. The impact of zero-shot CoT varies, with gains in mathematics but overall weaker improvements compared to English.}
	\label{fig:gpt3J}
\end{figure}

\begin{figure}[p]
  \centering
        \includegraphics[width=1.0\textwidth]{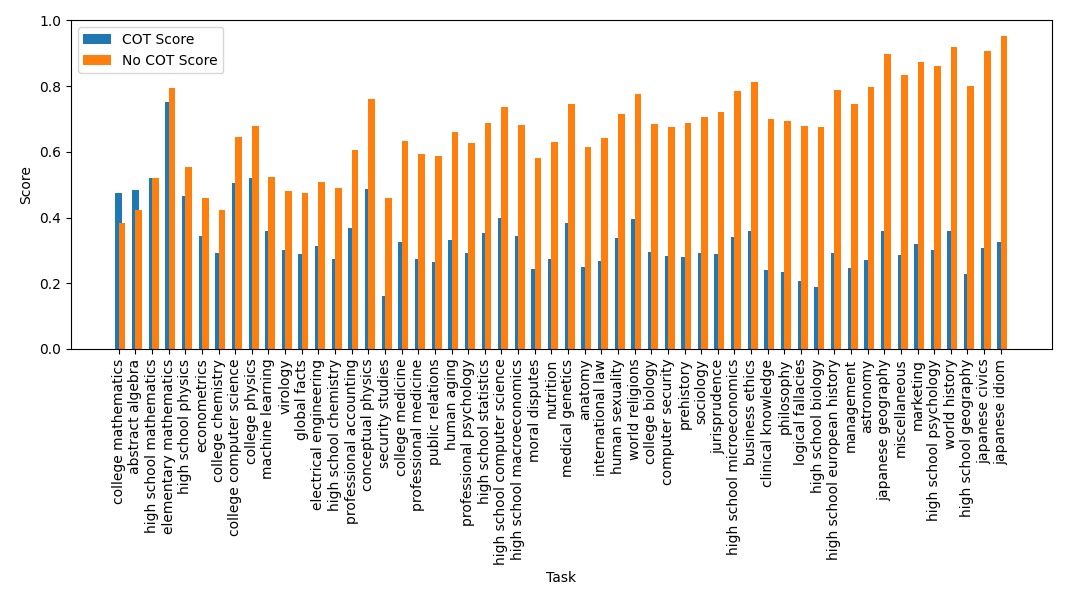}
	\caption{Performance across subjects for GPT-4o-mini in Japanese. While the overall trend remains negative, the decline is less severe compared to English, with a few subject areas benefiting from CoT.}
	\label{fig:gpt4J}
\end{figure}

For a more fine-grained perspective on how tasks in different subject areas respond to CoT, the four graphs of Figures~\ref{fig:gpt3E} to ~\ref{fig:gpt4J} show the detailed results for both English and Japanese. In these graphs, ``CoT Score'' indicates the performance when using CoT prompting, and the subject areas are arranged in descending order of score improvements due to CoT.

With the exception of GPT-4o in English, each dataset has some individual task areas that show improvement with CoT, despite the overall results from Table~\ref{tab:overall} that the overall performance change remains negative. This reinforces the impression of CoT as a task-dependent technique. 

The following section explores these results in more detail.

\ \\
\ \\


\section{Discussion}

\subsection*{Performance Differences in GPT-3.5}

Our experiments with GPT-3.5 showed that \textbf{zero-shot CoT resulted in lower overall scores} in both English and Japanese. However, the performance decline was more pronounced in Japanese (\(-0.059\)) compared to English (\(-0.022\)). Additionally, \textbf{CoT improved performance in 13 tasks for English} but only \textbf{8 tasks for Japanese}, suggesting that CoT was more beneficial in English.

A statistical analysis of the results showed that for \textbf{Japanese (JMMLU)}, CoT had a \textbf{significant effect} (\(t = 2.29, p = 0.02\)), whereas for \textbf{English (MMLU)}, the difference was \textbf{not statistically significant} (\(t = 0.79, p = 0.46\)). This suggests that in Japanese, \textbf{CoT significantly impacted task performance}, whereas in English, the effect was more variable.

\subsection*{Task-Level Performance Trends}

Examining individual tasks, \textbf{elementary mathematics} saw the \textbf{largest improvement due to CoT} in both languages (\(+0.247\) in Japanese, \(+0.278\) in English), indicating a similar effect. However, \textbf{clinical knowledge} showed a \textbf{strong contrast}: in English, \textbf{CoT improved performance} (\(+0.328\)), while in Japanese, it \textbf{decreased scores} (\(-0.120\)).

The \textbf{tasks most negatively affected by CoT} differed between languages:
\begin{itemize}
    \item \textbf{Japanese}: The largest declines occurred in \textit{international law, security studies, jurisprudence, moral disputes, and virology}.
    \item \textbf{English}: The most affected tasks were \textit{virology, high school U.S. history, high school geography, high school government and politics, and college biology}.
    \item The \textbf{greatest single drop} was \(-0.217\) in Japanese and \(-0.178\) in English.
\end{itemize}

\subsection*{Performance Differences in GPT-4o}

In contrast to GPT-3.5, \textbf{GPT-4o exhibited a more severe performance drop due to CoT}. Interestingly, the trend was reversed:
\begin{itemize}
    \item In \textbf{GPT-3.5}, CoT caused a \textbf{greater decline in Japanese} than in English.
    \item In \textbf{GPT-4o}, CoT resulted in a \textbf{larger drop in English}, while \textbf{Japanese scores remained more stable}.
    \item \textbf{Without CoT}, GPT-4o’s \textbf{Japanese performance increased significantly}, bringing it \textbf{closer to English-level scores}.
\end{itemize}

A statistical analysis of GPT-4o’s performance showed \textbf{significant differences in both languages}, with results of \textbf{JMMLU} (\(t = 17.0, p = 3.61 \times 10^{-25}\)) and \textbf{MMLU} (\(t = 14.1, p = 1.60 \times 10^{-25}\)). These results indicate that CoT had a much stronger impact on GPT-4o than GPT-3.5.

The number of tasks where CoT improved scores was minimal: only \textbf{two in JMMLU}, and in MMLU, \textbf{no tasks showed an increase}. The JMMLU tasks showing improvements were:
\begin{itemize}
    \item \textbf{College mathematics} (+0.091)
    \item \textbf{Abstract algebra} (+0.061)
\end{itemize}

For comparison, in GPT-3.5, these tasks showed a decrease in \textbf{college mathematics} (\(-0.040\)) and an increase in \textbf{abstract algebra} (\(+0.090\)).

The largest decrease was \textbf{Japanese idioms}\textit{} (\(-0.626\)), a Japan-specific category absent from MMLU. The limited number of Japanese-specific training samples may be a factor, though further analysis is needed to confirm this.

\subsection*{Understanding Why GPT-4o Performs Differently}

Although LLMs function as a black box, differences in generated outputs suggest that GPT-3.5 and GPT-4o approach reasoning differently.  In the no-CoT setting, GPT-3.5’s responses often contained fewer \textbf{reasoning steps}, whereas GPT-4o already exhibited more detailed reasoning even without explicit CoT prompting.

One viable interpretation is that for GPT-3.5, explicit CoT prompting can increase the number of reasoning steps, encouraging a \textbf{``question → CoT → answer''} sequence that improves accuracy in mathematics and reasoning-related tasks. However, if GPT-4o already incorporates CoT-like reasoning internally, additional CoT prompting can become redundant or even disruptive. This would align with recent advancements in models like DeepSeek-R1, which perform well on CoT-style tasks and may have benefited from reasoning-based instruction tuning~\cite{deepseekai2025}.  

A related factor that may contribute to performance decline is the \textbf{``increase in clauses'' effect} discussed by~\cite{mirzadeh2024gsm}. Their analysis of Putnam mathematical problems found that longer clause structures correlate with performance degradation. Thus, the increased clause length resulting from adding a CoT phrase such as ``Let's think step by step'' may introduce complexity that outweighs any potential benefits in reasoning.

\subsection*{Overall Comparison Between English and Japanese}

Finally, comparing results across the two languages:
\begin{itemize}
    \item For \textbf{GPT-3.5}, \textbf{CoT was more effective in English}, with more tasks showing score increases.
    \item For \textbf{GPT-4o}, \textbf{CoT was more effective in Japanese}, with a \textbf{greater performance decline in English}.
    \item The findings suggest that as \textbf{LLMs evolve}, their reliance on explicit CoT prompting \textbf{changes}, depending on how much reasoning is already embedded in their responses.
\end{itemize}

\section{Conclusions}

This study examined the impact of zero-shot Chain-of-Thought (CoT) prompting in Japanese and English tasks using GPT-3.5 and GPT-4o-mini. For GPT-3.5, CoT led to an overall decline in performance, while showing some benefits in reasoning-based tasks like mathematics. For GPT-4o-mini, CoT resulted in significant performance drops in both languages, with only a few Japanese task categories showing improvements.

The underlying mechanisms of CoT remain incompletely understood, and optimising its application relies in part on empirical testing. Our cross-language study identifies some inconsistencies in CoT’s impact across different application areas, underscoring the broader challenge of interpreting how LLMs process reasoning prompts. As models continue to evolve, bridging this knowledge gap will be essential for developing a clearer picture of how internal LLM reasoning structures emerge—whether through explicit prompting or as an inherent part of next-generation architectures.

\subsection*{Acknowledgements}
Generative AI tools were used to assist in the drafting of the paper and in the generation of English translations. In all cases, the responsibility for the accuracy and completeness and style of the final text lies with the authors.

\bibliographystyle{unsrtnat}
\bibliography{main}

\begin{thebibliography}{12}
\providecommand{\natexlab}[1]{#1}
\providecommand{\url}[1]{\texttt{#1}}
\expandafter\ifx\csname urlstyle\endcsname\relax
  \providecommand{\doi}[1]{doi: #1}\else
  \providecommand{\doi}{doi: \begingroup \urlstyle{rm}\Url}\fi

\bibitem[Radford et~al.(2019)Radford, Wu, Child, Luan, Amodei, and
  Sutskever]{radford2019gpt2}
Alec Radford, Jeffrey Wu, Rewon Child, David Luan, Dario Amodei, and Ilya
  Sutskever.
\newblock Language models are unsupervised multitask learners, 2019.
\newblock URL
  \url{https://cdn.openai.com/better-language-models/language_models_are_unsupervised_multitask_learners.pdf}.
\newblock Technical report.

\bibitem[Brown et~al.(2020)Brown, Mann, Ryder, Subbiah, Kaplan, Dhariwal,
  Neelakantan, Shyam, Sastry, Askell, Agarwal, Herbert-Voss, Krueger, Henighan,
  Child, Ramesh, Ziegler, Wu, Winter, Hesse, Chen, Sigler, Litwin, Gray, Chess,
  Clark, Berner, McCandlish, Radford, Sutskever, and Amodei]{brown2020fewshot}
Tom~B. Brown, Benjamin Mann, Nick Ryder, Melanie Subbiah, Jared Kaplan,
  Prafulla Dhariwal, Arvind Neelakantan, Pranav Shyam, Girish Sastry, Amanda
  Askell, Sandhini Agarwal, Ariel Herbert-Voss, Gretchen Krueger, Tom Henighan,
  Rewon Child, Aditya Ramesh, Daniel~M. Ziegler, Jeffrey Wu, Clemens Winter,
  Christopher Hesse, Mark Chen, Eric Sigler, Mateusz Litwin, Scott Gray,
  Benjamin Chess, Jack Clark, Christopher Berner, Sam McCandlish, Alec Radford,
  Ilya Sutskever, and Dario Amodei.
\newblock Language models are few-shot learners, 2020.
\newblock URL \url{https://arxiv.org/abs/2005.14165}.

\bibitem[Hendrycks et~al.(2021)Hendrycks, Burns, Basart, Zou, Mazeika, Song,
  and Steinhardt]{hendrycks2021MMLU}
Dan Hendrycks, Collin Burns, Steven Basart, Andy Zou, Mantas Mazeika, Dawn
  Song, and Jacob Steinhardt.
\newblock Measuring massive multitask language understanding, 2021.
\newblock URL \url{https://arxiv.org/abs/2009.03300}.

\bibitem[Yamada et~al.(2023)]{jmmlu2023}
Ikuya Yamada et~al.
\newblock Japanese multi-task language understanding benchmark ({JMMLU}), 2023.
\newblock URL \url{https://huggingface.co/datasets/nlp-waseda/JMMLU}.
\newblock Accessed: 2025-03-08.

\bibitem[Wei et~al.(2022)Wei, Wang, Schuurmans, Bosma, Ichter, Xia, Chi, Le,
  and Zhou]{wei2022chain}
Jason Wei, Xuezhi Wang, Dale Schuurmans, Maarten Bosma, Brian Ichter, Fei Xia,
  Ed~H. Chi, Quoc~V. Le, and Denny Zhou.
\newblock Chain-of-thought prompting elicits reasoning in large language
  models.
\newblock In \emph{Proceedings of the 36th International Conference on Neural
  Information Processing Systems}, NIPS'22, pages 24824--24837. Curran
  Associates Inc., 2022.
\newblock URL \url{https://dl.acm.org/doi/10.5555/3600270.3602070}.

\bibitem[Kojima et~al.(2023)Kojima, Gu, Reid, Matsuo, and
  Iwasawa]{kojima2023large}
Takeshi Kojima, Shixiang~Shane Gu, Machel Reid, Yutaka Matsuo, and Yusuke
  Iwasawa.
\newblock Large language models are zero-shot reasoners, 2023.
\newblock URL \url{https://arxiv.org/abs/2205.11916}.

\bibitem[Jin et~al.(2024)Jin, Yu, Shu, Zhao, Hua, Meng, Zhang, and
  Du]{jin2024impact}
Mingyu Jin, Qinkai Yu, Dong Shu, Haiyan Zhao, Wenyue Hua, Yanda Meng, Yongfeng
  Zhang, and Mengnan Du.
\newblock The impact of reasoning step length on large language models, 2024.
\newblock URL \url{https://arxiv.org/abs/2401.04925}.

\bibitem[Mirzadeh et~al.(2024)Mirzadeh, Alizadeh, Shahrokhi, Tuzel, Bengio, and
  Farajtabar]{mirzadeh2024gsm}
Iman Mirzadeh, Keivan Alizadeh, Hooman Shahrokhi, Oncel Tuzel, Samy Bengio, and
  Mehrdad Farajtabar.
\newblock {GSM}-{S}ymbolic: {U}nderstanding the limitations of mathematical
  reasoning in large language models, 2024.
\newblock URL \url{https://arxiv.org/abs/2410.05229}.

\bibitem[Gulati et~al.(2024)Gulati, Miranda, Chen, Xia, Fronsdal,
  de~Moraes~Dumont, and Koyejo]{gulati2024putnam}
Aryan Gulati, Brando Miranda, Eric Chen, Emily Xia, Kai Fronsdal, Bruno
  de~Moraes~Dumont, and Sanmi Koyejo.
\newblock Putnam-{AXIOM}: A functional and static benchmark for measuring
  higher level mathematical reasoning.
\newblock In \emph{The 4th Workshop on Mathematical Reasoning and AI at
  NeurIPS'24}, 2024.
\newblock URL \url{https://openreview.net/forum?id=YXnwlZe0yf}.

\bibitem[Yin et~al.(2024)Yin, Wang, Horio, Kawahara, and Sekine]{yin2024should}
Ziqi Yin, Hao Wang, Kaito Horio, Daisuke Kawahara, and Satoshi Sekine.
\newblock Should we respect {LLMs}? {A} cross-lingual study on the influence of
  prompt politeness on {LLM} performance, 2024.
\newblock URL \url{https://arxiv.org/abs/2402.14531}.

\bibitem[Li et~al.(2023)Li, Wang, Zhang, Zhu, Hou, Lian, Luo, Yang, and
  Xie]{li2023large}
Cheng Li, Jindong Wang, Yixuan Zhang, Kaijie Zhu, Wenxin Hou, Jianxun Lian,
  Fang Luo, Qiang Yang, and Xing Xie.
\newblock Large language models understand and can be enhanced by emotional
  stimuli, 2023.
\newblock URL \url{https://arxiv.org/abs/2307.11760}.

\bibitem[DeepSeek-AI et~al.(2025)DeepSeek-AI, Guo, Yang,
  et~al.]{deepseekai2025}
DeepSeek-AI, Daya Guo, Dejian Yang, et~al.
\newblock Deep{S}eek-{R}1: Incentivizing reasoning capability in {LLM}s via
  reinforcement learning, 2025.
\newblock URL \url{https://arxiv.org/abs/2501.12948}.

\end{thebibliography}

\end{document}